\documentclass[10pt,twocolumn,letterpaper]{article}

\usepackage{iccv}
\usepackage{times}
\usepackage{epsfig}
\usepackage{graphicx}
\usepackage{amsmath}
\usepackage{amssymb}

\usepackage{siunitx}
\usepackage{cite}
\usepackage{ctable}
\usepackage{hhline}
\usepackage{booktabs}
\usepackage{subcaption}
\usepackage{csquotes}

\newcolumntype{L}[1]{>{\raggedright\let\newline\\\arraybackslash\hspace{0pt}}m{#1}}
\newcolumntype{C}[1]{>{\centering\let\newline\\\arraybackslash\hspace{0pt}}m{#1}}
\newcolumntype{R}[1]{>{\raggedleft\let\newline\\\arraybackslash\hspace{0pt}}m{#1}}

\usepackage[breaklinks=true,bookmarks=false]{hyperref}

\iccvfinalcopy 

\def\iccvPaperID{****} 

\ificcvfinal\fi

\title{Camera Distance-aware Top-down Approach for 3D Multi-person Pose Estimation from a Single RGB Image}

\author{
Gyeongsik Moon$^1$\hspace{1.0cm} Ju Yong Chang$^2$ \hspace{1.0cm} Kyoung Mu Lee$^1$\\
\\
\hspace{-0.9cm}
$^1$ECE \& ASRI, Seoul National University, Korea \hspace{1.0cm}
$^2$ECE, Kwangwoon University, Korea \\
{\small \texttt {\{mks0601, kyoungmu\}@snu.ac.kr}, \texttt {juyong.chang@gmail.com}}
}

\begin{document}

\twocolumn[{
\maketitle
\vspace{-2.5em}
\centerline{
\includegraphics[width=\linewidth,trim={4pt 4pt 4pt 4pt}]{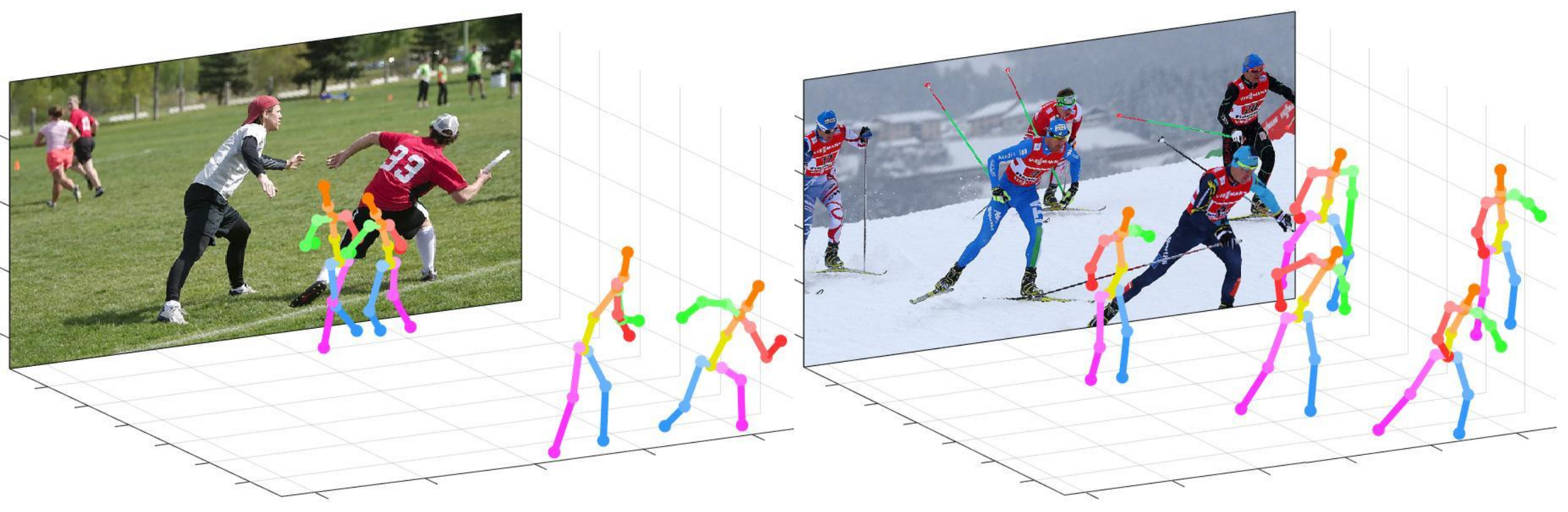}}
\captionof{figure}
{
Qualitative results of applying our 3D multi-person pose estimation framework to COCO dataset~\cite{lin2014microsoft} which consists of \emph{in-the-wild} images. Most of the previous 3D human pose estimation studies mainly focused on the root-relative 3D single-person pose estimation. In this study, we propose a general 3D \emph{multi}-person pose estimation framework that takes into account all factors including human detection and 3D human root localization.
}
\label{fig:qualitative}
\vspace{1em}
}]

\maketitle
\thispagestyle{empty}

\begin{abstract}
Although significant improvement has been achieved recently in 3D human pose estimation, most of the previous methods only treat a single-person case. 
In this work, we firstly propose a fully learning-based, camera distance-aware top-down approach for 3D multi-person pose estimation from a single RGB image. 
The pipeline of the proposed system consists of human detection, absolute 3D human root localization, and root-relative 3D single-person pose estimation modules.
Our system achieves comparable results with the state-of-the-art 3D single-person pose estimation models without any groundtruth information and significantly outperforms previous 3D multi-person pose estimation methods on publicly available datasets. The code is available in \footnote{\url{https://github.com/mks0601/3DMPPE_ROOTNET_RELEASE}}\textsuperscript{,}\footnote{\url{https://github.com/mks0601/3DMPPE_POSENET_RELEASE}}.
\end{abstract}

\section{Introduction}

The goal of 3D human pose estimation is to localize semantic keypoints of single or multiple human bodies in 3D space. 
It is an essential technique for human behavior understanding and human-computer interaction. 
Recently, many methods~\cite{pavlakos2017coarse,sun2017compositional,zhou2017weaklysupervised,martinez2017simple,yang20183d,sun2018integral} utilize deep convolutional neural networks (CNNs) and have achieved noticeable performance improvement on large-scale publicly available datasets~\cite{ionescu2014human3,mehta2017monocular}.

Most of the previous 3D human pose estimation methods~\cite{pavlakos2017coarse,sun2017compositional,zhou2017weaklysupervised,martinez2017simple,yang20183d,sun2018integral} are designed for single-person case. 
They crop the human area in an input image with a groundtruth bounding box or the bounding box that is predicted from a human detection model~\cite{he2017mask}. 
The cropped patch of a human body is fed into the 3D pose estimation module, which then estimates the 3D location of each keypoint. As their models take a single cropped image, estimating the absolute camera-centered coordinate of each keypoint is difficult. 
To handle this issue, many methods~\cite{pavlakos2017coarse,sun2017compositional,zhou2017weaklysupervised,martinez2017simple,yang20183d,sun2018integral} estimate the relative 3D pose to a reference point in the body, e.g., the center joint (\textit{i.e.}, pelvis) of a human, called {\it root}. 
The final 3D pose is obtained by adding the 3D coordinates of the root to the estimated root-relative 3D pose. 
Prior information on the bone length~\cite{pavlakos2017coarse} or the groundtruth~\cite{sun2018integral} has been commonly used for the localization of the root.

Recently, many top-down approaches~\cite{huang2017coarse,chen2018cascaded,xiao2018simple} for the 2D multi-person pose estimation have shown noticeable performance improvement. 
These approaches first detect humans by using a human detection module, and then estimate the 2D pose of each human by a 2D single-person pose estimation module. 
Although they are straightforward when used in 2D cases, extending them to 3D cases is nontrivial. 
Note that for the estimation of 3D multi-person poses, we need to know the absolute distance to each human from the camera as well as the 2D bounding boxes. 
However, existing human detectors provide 2D bounding boxes only. 

\begin{figure*}
\begin{center}
\includegraphics[width=1.0\linewidth]{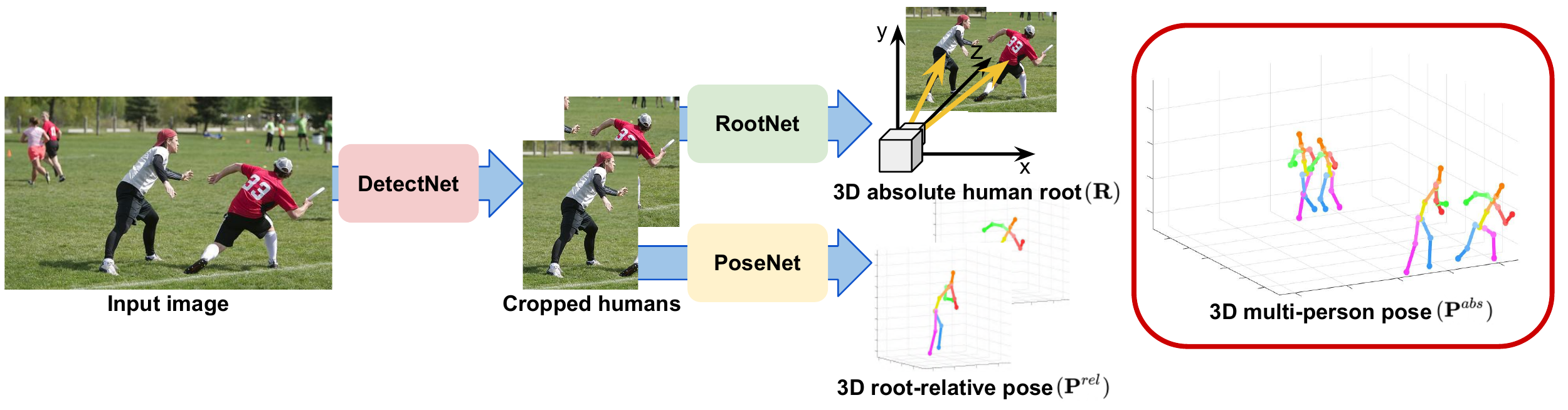}
\end{center}
\vspace*{-5mm}
   \caption{Overall pipeline of the proposed framework for 3D multi-person pose estimation from a single RGB image. The proposed framework can recover the absolute camera-centered coordinates of multiple persons' keypoints.}
\vspace*{-3mm}
\label{fig:overall_pipeline}
\end{figure*}

In this study, we propose a general framework for 3D multi-person pose estimation. 
To the best of our knowledge, this study is the first to propose a fully learning-based camera distance-aware top-down approach of which components are compatible with most of the previous human detection and 3D human pose estimation methods. 
The pipeline of the proposed system consists of three modules. First, a human detection network (DetectNet) detects the bounding boxes of humans in an input image. 
Second, the proposed 3D human root localization network (RootNet) estimates the camera-centered coordinates of the detected humans' roots. Third, a root-relative 3D single-person pose estimation network (PoseNet) estimates the root-relative 3D pose for each detected human. Figures~\ref{fig:qualitative} and~\ref{fig:overall_pipeline} show the qualitative results and overall pipeline of our framework, respectively.

We show that our approach outperforms previous 3D multi-person pose estimation methods~\cite{rogez2017lcr,mehta2018single} on several publicly available 3D single- and multi-person pose estimation datasets~\cite{ionescu2014human3,mehta2018single} by a large margin. Also, even without any groundtruth information (\textit{i.e.}, the bounding boxes and the 3D location of the roots), our method achieves comparable performance with the state-of-the-art 3D single-person pose estimation methods that use the groundtruth in the inference time. 
Note that our framework is new but follows previous conventions of object detection and 3D human pose estimation networks. 
Thus, previous detection and pose estimation methods can be easily plugged into our framework, which makes the proposed framework quite flexible and generalizable. 

Our contributions can be summarized as follows.

\begin{itemize}
\item We propose a new general framework for 3D multi-person pose estimation from a single RGB image. 
The framework is the first fully learning-based, camera distance-aware top-down approach, of which components are compatible with most of the previous human detection and 3D human pose estimation models.

\item Our framework outputs the absolute camera-centered coordinates of multiple humans' keypoints. 
For this, we propose a 3D human root localization network (RootNet). 
This model makes it easy to extend the 3D single-person pose estimation techniques to the absolute 3D pose estimation of multiple persons.

\item We show that our method significantly outperforms previous 3D multi-person pose estimation methods on several publicly available datasets. 
Also, it achieves comparable performance with the state-of-the-art 3D single-person pose estimation methods without any groundtruth information.
\end{itemize}

\section{Related works}

\textbf{2D multi-person pose estimation.}
There are two main approaches in the multi-person pose estimation. 
The first one, top-down approach, deploys a human detector that estimates the bounding boxes of humans. 
Each detected human area is cropped and fed into the pose estimation network. 
The second one, bottom-up approach, localizes all human body keypoints in an input image first, and then groups them into each person using some clustering techniques. 

~\cite{papandreou2017towards,huang2017coarse,chen2018cascaded,xiao2018simple,moon2019posefix,moon2019multi} are based on the top-down approach. Papandreou~\etal~\cite{papandreou2017towards} predicted 2D offset vectors and 2D heatmaps for each joint. 
They fused the estimated vectors and heatmaps to generate highly localized heatmaps. Chen~\etal~\cite{chen2018cascaded} proposed a cascaded pyramid network whose cascaded structure refines an initially estimated pose by focusing on hard keypoints. Xiao~\etal~\cite{xiao2018simple} used a simple pose estimation network that consists of a deep backbone network and several upsampling layers.

~\cite{pishchulin2016deepcut,insafutdinov2016deepercut,cao2016realtime,newell2017associative,kocabas2018multiposenet} are based on the bottom-up approach. Cao~\etal~\cite{cao2016realtime} proposed the part affinity fields (PAFs) that model the association between human body keypoints. 
They grouped the localized keypoints of all persons in the input image by using the estimated PAFs. Newell~\etal~\cite{newell2017associative} introduced a pixel-wise tag value to assign localized keypoints to a certain human. Kocabas~\etal~\cite{kocabas2018multiposenet} proposed a pose residual network for assigning detected keypoints to each person.

\textbf{3D single-person pose estimation.}
Current 3D single-person pose estimation methods can be categorized into single- and two-stage approaches. 
The single-stage approach directly localizes the 3D body keypoints from the input image. 
The two-stage methods utilize the high accuracy of 2D human pose estimation. They initially localize body keypoints in a 2D space and lift them to a 3D space.

~\cite{li20143d,tekin2016structured,pavlakos2017coarse,sun2017compositional,sun2018integral} are based on the single-stage approach. 
Li~\etal~\cite{li20143d} proposed a multi-task framework that jointly trains both the pose regression and body part detectors. Tekin~\etal~\cite{tekin2016structured} modeled high-dimensional joint dependencies by adopting an auto-encoder structure. 
Pavlakos~\etal~\cite{pavlakos2017coarse} extended the U-net shaped network to estimate a 3D heatmap for each joint. They used a coarse-to-fine approach to boost performance. Sun~\etal~\cite{sun2017compositional} introduced compositional loss to consider the joint connection structure. 
Sun~\etal~\cite{sun2018integral} used soft-argmax operation to obtain the 3D coordinates of body joints in a differentiable manner.

~\cite{park20163d,chen20173d,martinez2017simple,zhou2017weaklysupervised,fang2018learning,yang20183d,chang20182d} are based on the two-stage approach. Park~\etal~\cite{park20163d} estimated the initial 2D pose and utilized it to regress the 3D pose. Martinez~\etal~\cite{martinez2017simple} proposed a simple network that directly regresses the 3D coordinates of body joints from 2D coordinates. Zhou~\etal~\cite{zhou2017weaklysupervised} proposed a geometric loss to facilitate weakly supervised learning of the depth regression module with images in the wild. 
Yang~\etal~\cite{yang20183d} utilized adversarial loss to handle the 3D human pose estimation in the wild.

\textbf{3D multi-person pose estimation.}
Few studies have been conducted on 3D multi-person pose estimation from a single RGB image. Rogez~\etal~\cite{rogez2017lcr} proposed a top-down approach called LCR-Net, which consists of localization, classification, and regression parts.
The localization part detects a human from an input image, and the classification part classifies the detected human into several anchor-poses.
The anchor-pose is defined as a pair of 2D and root-relative 3D pose. It is generated by clustering poses in the training set.
Then, the regression part refines the anchor-poses. Mehta~\etal~\cite{mehta2018single} proposed a bottom-up approach system.
They introduced an occlusion-robust pose-map formulation which supports pose inference for more than one person through PAFs~\cite{cao2016realtime}.

\textbf{3D human root localization in 3D multi-person pose estimation.}
Rogez~\etal~\cite{rogez2017lcr} estimated both the 2D pose in the image coordinate space and the 3D pose in the camera-centered coordinate space simultaneously. 
They obtained the 3D location of the human root by minimizing the distance between the estimated 2D pose and projected 3D pose, similar to what Mehta~\etal~\cite{mehta2017monocular} did. 
However, this strategy cannot be generalized to other 3D human pose estimation methods because it requires both the 2D and 3D estimations. For example, many works~\cite{sun2018integral,zhou2017weaklysupervised,pavlakos2017coarse,yang20183d} estimate the 2D image coordinates and root-relative depth values of keypoints. 
As their methods do not output root-relative camera-centered coordinates of keypoints, such a distance minimization strategy cannot be used. 
Moreover, contextual information cannot be exploited because the image feature is not considered. For example, it cannot distinguish between a child close to the camera and an adult far from the camera because their scales in the 2D image is similar.

\section{Overview of the proposed model}

The goal of our system is to recover the absolute camera-centered coordinates of multiple persons' keypoints $\{\mathbf{P}^{abs}_j\}_{j=1}^J$, where $J$ denotes the number of joints. To address this problem, we construct our system based on the top-down approach that consists of DetectNet, RootNet, and PoseNet. 
The DetectNet detects a human bounding box of each person in the input image. 
The RootNet takes the cropped human image from the DetectNet and localizes the root of the human $\mathbf{R}=(x_R,y_R,Z_R)$, in which $x_R$ and $y_R$ are pixel coordinates, and $Z_R$ is an absolute depth value. 
The same cropped human image is fed to the PoseNet, which estimates the root-relative 3D pose $\mathbf{P}^{rel}_{j}=(x_j,y_j,Z^{rel}_j)$, in which $x_j$ and $y_j$ are pixel coordinates in the cropped image space and $Z^{rel}_j$ is root-relative depth value. 
We convert $Z^{rel}_j$ into $Z^{abs}_j$ by adding $Z_R$ and transform $x_j$ and $y_j$ to the original input image space. 
Then, the final absolute 3D pose $\{\mathbf{P}^{abs}_j\}_{j=1}^J$ is obtained by simple back-projection.


\section{DetectNet}

We use Mask R-CNN~\cite{he2017mask} as the framework of DetectNet. 
Mask R-CNN~\cite{he2017mask} consists of three parts. 
The first one, backbone, extracts useful local and global features from the input image by using deep residual network (ResNet)~\cite{he2016deep} and feature pyramid network~\cite{lin2017feature}. 
Based on the extracted features, the second part, region proposal network, proposes human bounding box candidates. 
The RoIAlign layer extracts the features of each proposal and passes them to the third part, which is the classification head network. 
The head network determines whether the given proposal is a human or not and estimates the bounding box refinement offsets. 
It achieves the state-of-the-art performance on publicly available object detection datasets~\cite{lin2014microsoft}. 
Due to its high performance and publicly available code~\cite{Detectron2018,massa2018mrcnn}, we use Mask R-CNN~\cite{he2017mask} as a DetectNet in our pipeline.

\begin{figure}[t]
\begin{center}
   \includegraphics[width=1.0\linewidth]{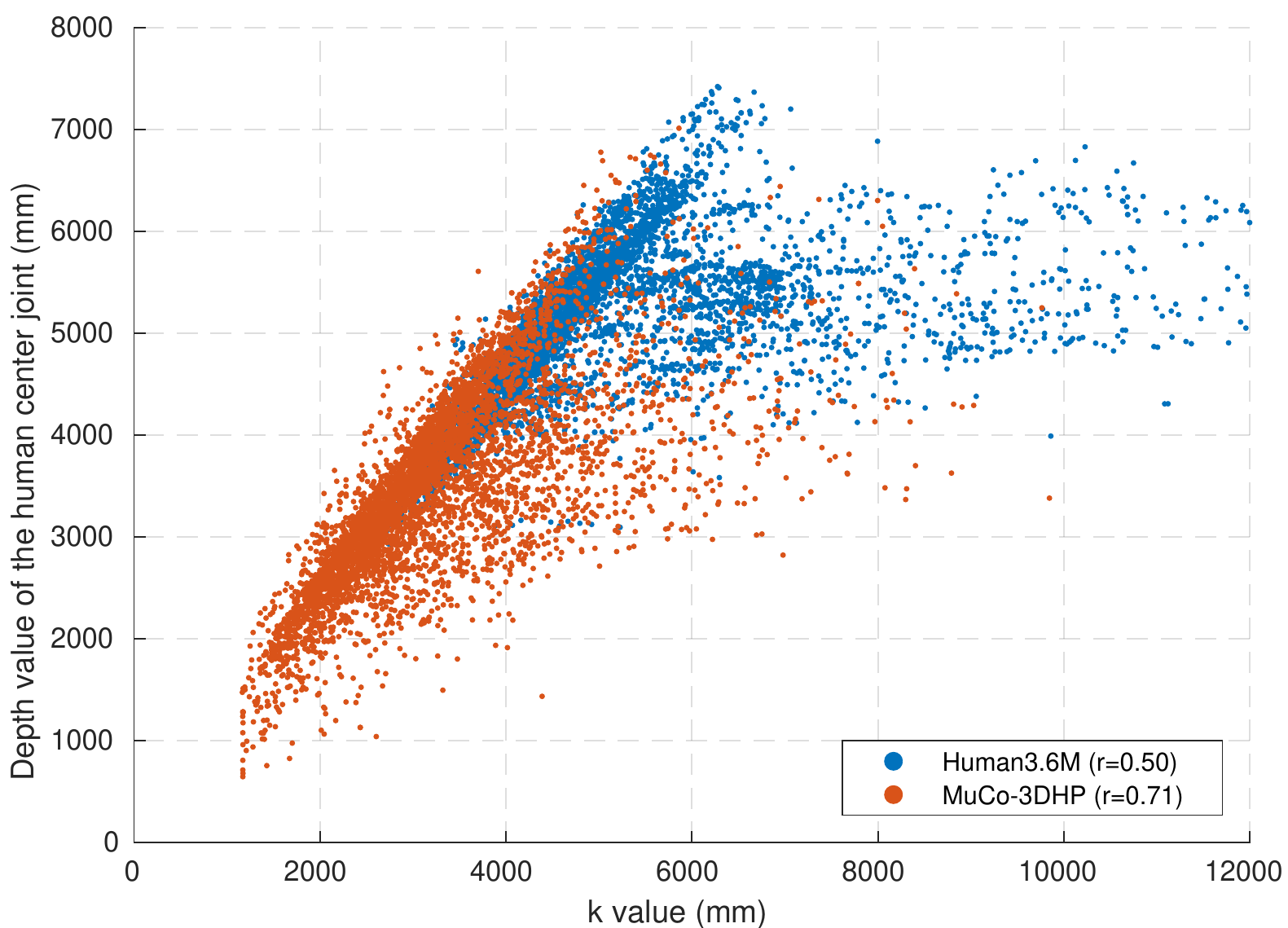}
\end{center}
\vspace*{-5mm}
   \caption{Correlation between $k$ and real depth value of the human root. Human3.6M~\cite{ionescu2014human3} and MuCo-3DHP~\cite{mehta2018single} datasets were used. $r$ represents Pearson correlation coefficient.}
\vspace*{-3mm}
\label{fig:k_vs_distance}
\end{figure}

\section{RootNet}

\subsection{Model design}
The RootNet estimates the camera-centered coordinates of the human root $\mathbf{R}=(x_R,y_R,Z_R)$ from a cropped human image. 
To obtain them, RootNet separately estimates the 2D image coordinates $(x_R,y_R)$ and the depth value (\textit{i.e.}, the distance from the camera $Z_R$) of the human root. 
The estimated 2D image coordinates are back-projected to the camera-centered coordinate space using the estimated depth value, which becomes the final output.

Considering that an image provides sufficient information on where the human root is located in the image space, the 2D estimation part can learn to localize it easily. 
By contrast, estimating the depth only from a cropped human image is difficult because the input does not provide information on the relative position of the camera and human. 
To resolve this issue, we introduce a new distance measure, $k$, which is defined as follows:
\begin{equation}
k=\sqrt{\alpha_x\alpha_y\frac{A_{real}}{A_{img}}},
\label{eq:k}
\end{equation}
where $\alpha_x$, $\alpha_y$, $A_{real}$, and $A_{img}$ are focal lengths divided by the per-pixel distance factors (pixel) of $x$- and $y$-axes, the area of the human in real space ($mm^2$), and image space (pixel$^2$), respectively. 
$k$ approximates the absolute depth from the camera to the object using the ratio of the actual area and the imaged area of it, given camera parameters. 
Eq~\ref{eq:k} can be easily derived by considering a pinhole camera projection model. 
The distance $d$ ($mm$) between the camera and object can be calculated as follows:
\begin{eqnarray}
d=\alpha_x\frac{l_{x,real}}{l_{x,img}} = \alpha_y\frac{l_{y,real}}{l_{y,img}},
\label{eq:d}
\end{eqnarray}
where $l_{x,real}$, $l_{x,img}$, $l_{y,real}$, $l_{y,img}$ are the lengths of an object in real space ($mm$) and in image space (pixel), on the $x$ and $y$-axes, respectively. 
By multiplying the two representations of $d$ in Eq~\ref{eq:d} and taking the square root of it, we can have the 2D extended version of depth measure $k$ in Eq~\ref{eq:k}. 
Assuming that $A_{real}$ is constant and using $\alpha_x$ and $\alpha_y$ from datasets, the distance between the camera and an object can be measured from the area of the bounding box. 
As we only consider humans, we assume that $A_{real}$ is $2000mm \times 2000mm$. 
The area of the human bounding box is used as $A_{img}$ after extending it to fixed aspect ratio (\textit{i.e.}, height:width = 1:1). 
Figure~\ref{fig:k_vs_distance} shows that such an approximation provides a meaningful correlation between $k$ and the real depth values of the human root in 3D human pose estimation datasets~\cite{ionescu2014human3,mehta2018single}.

\begin{figure}[t]
\begin{center}
   \includegraphics[width=1.0\linewidth]{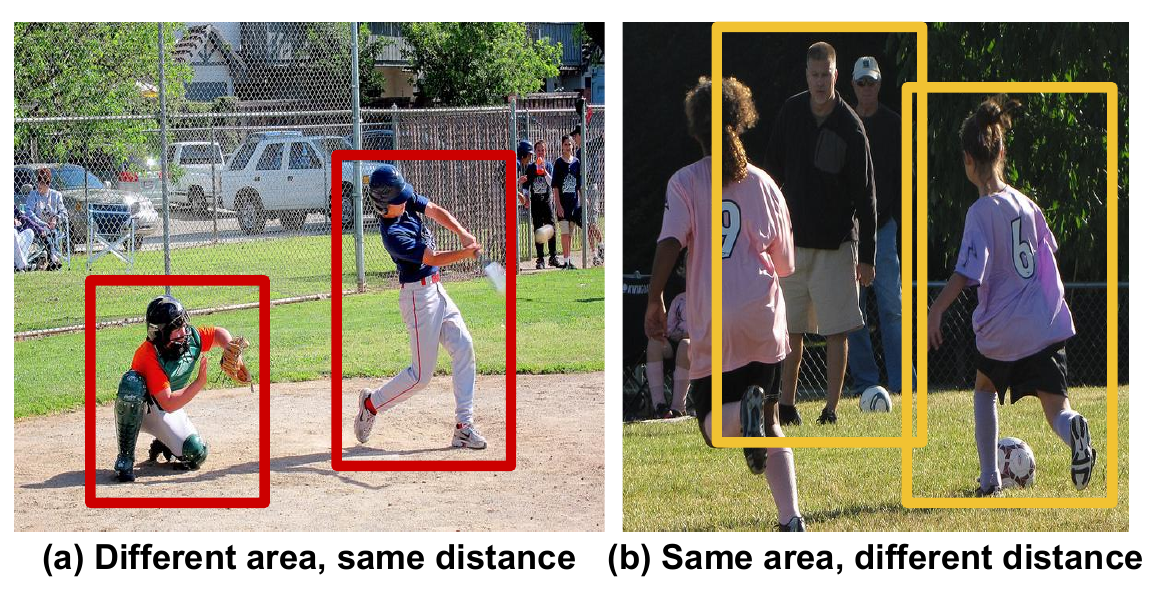}
\end{center}
\vspace*{-5mm}
   \caption{Examples where $k$ fails to represent the distance between a human and the camera because of incorrect $A_{img}$.}
\vspace*{-3mm}
\label{fig:wrong_Aimg_examples}
\end{figure}

Although $k$ can represent how far the human is from the camera, it can be wrong in several cases because it assumes that $A_{img}$ is an area of $A_{real}$ (\textit{i.e.}, $2000mm \times 2000mm$) in the image space when the distance between the human and the camera is $k$. 
However, as $A_{img}$ is obtained by extending the 2D bounding box, it can have a different value according to its appearance, although the distance to the camera is the same. 
For example, as shown in Figure~\ref{fig:wrong_Aimg_examples}(a), two humans have different $A_{img}$ although they are at the same distance to the camera. 
On the other hand, in some cases, $A_{img}$ can be the same, even with different distances from the camera. 
For example, in Figure~\ref{fig:wrong_Aimg_examples}(b), a child and an adult have similar $A_{img}$ however, the child is closer to the camera than the adult. 

To handle this issue, we design the RootNet to utilize the image feature to correct $A_{img}$, eventually $k$. 
The image feature can give a clue to the RootNet about how much the $A_{img}$ has to be changed. 
For example, in Figure~\ref{fig:wrong_Aimg_examples}(a), the left image can tell the RootNet to increase the area because the human is in a crouching posture. 
Also, in Figure~\ref{fig:wrong_Aimg_examples}(b), the right image can tell the RootNet to increase the area because the input image contains a child. 
Specifically, the RootNet outputs the correction factor $\gamma$ from the image feature. 
The estimated $\gamma$ is multiplied by the given $A_{img}$, which becomes $A_{img}^{\gamma}$. From $A_{img}^{\gamma}$, $k$ is calculated and it becomes the final depth value.

\begin{figure}[t]
\begin{center}
   \includegraphics[width=1.0\linewidth]{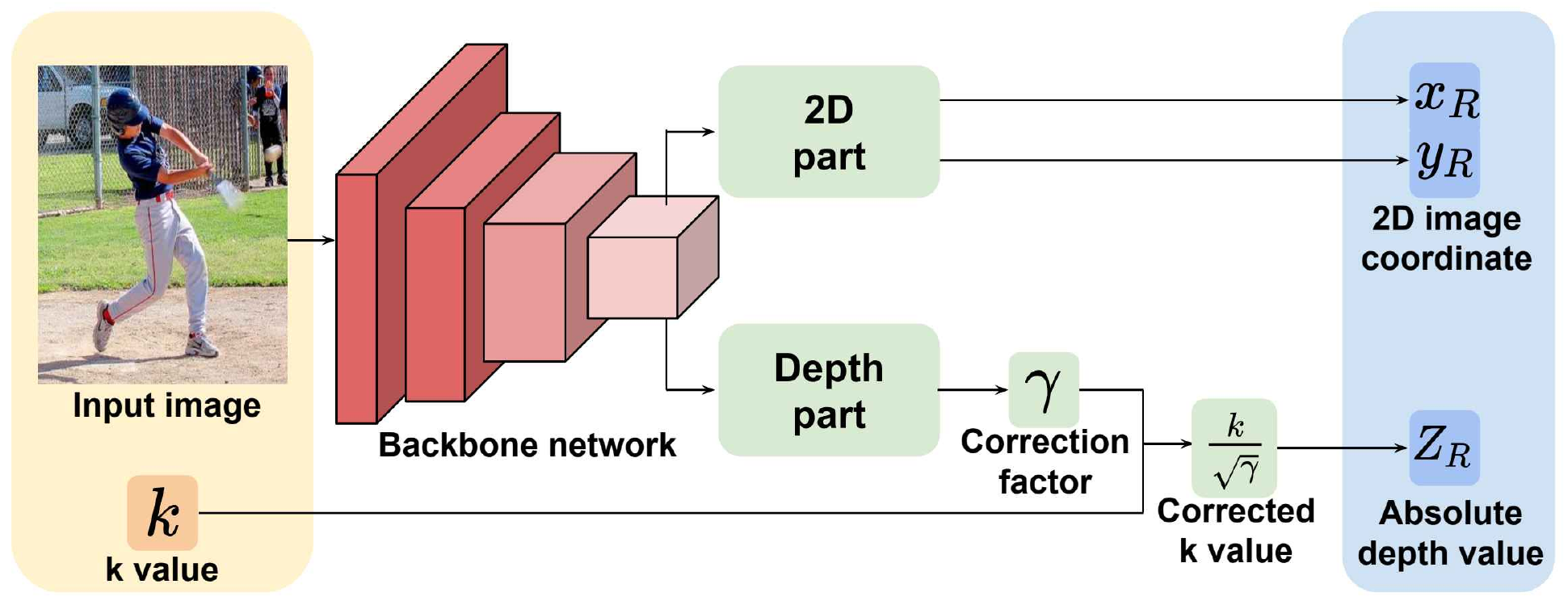}
\end{center}
\vspace*{-5mm}
   \caption{Network architecture of the RootNet. The RootNet estimates the 3D human root coordinate.}
\vspace*{-3mm}
\label{fig:rootnet_architecture}
\end{figure}

\subsection{Camera normalization}
Our RootNet outputs correction factor $\gamma$ only from an input image. Therefore, data from any camera intrinsic parameters (\textit{i.e.}, $\alpha_x$ and $\alpha_y$) can be used during training and testing. We call this property \textit{camera normalization}, which makes our RootNet very flexible. For example, in the training stage, data from different $\alpha_x$ and $\alpha_y$ can be used together. Also, in the testing stage, RootNet can be used when $\alpha_x$ and $\alpha_y$ are not available, likely for in-the-wild images. In this case, $\alpha_x$ and $\alpha_y$ can be set to any values $\alpha_x^\prime$ and $\alpha_y^\prime$, respectively. Then, estimated $Z_R$ represents distance between an object and camera whose $\alpha_x$ and $\alpha_y$ are $\alpha_x^\prime$ and $\alpha_y^\prime$, respectively. 

\subsection{Network architecture}
 The network architecture of the RootNet, which comprises three components, is visualized in Figure~\ref{fig:rootnet_architecture}. 
 First, a backbone network extracts the useful global feature of the input human image using ResNet~\cite{he2016deep}. 
 Second, the 2D image coordinate estimation part takes a feature map from the backbone part and upsamples it using three consecutive deconvolutional layers with batch normalization layers~\cite{ioffe2015batch} and ReLU activation function. 
 Then, a 1-by-1 convolution is applied to produce a 2D heatmap of the root. Soft-argmax~\cite{sun2018integral} extracts 2D image coordinates $x_R,y_R$ from the 2D heatmap. 
 The third component is the depth estimation part. It also takes a feature map from the backbone part and applies global average pooling. 
 Then, the pooled feature map goes through a 1-by-1 convolution, which outputs a single scalar value $\gamma$. 
 The final absolute depth value $Z_R$ is obtained by multiplying $k$ with $\frac{1}{\sqrt{\gamma}}$. 
 In practice, we implemented the RootNet to output $\gamma^{\prime}=\frac{1}{\sqrt{\gamma}}$ directly and multiply it with the $k$ to obtain the absolute depth value $Z_R$ (\textit{i.e.}, $Z_R=\gamma^{\prime}k$).

\subsection{Loss function}
We train the RootNet by minimizing the $L1$ distance between the estimated and groundtruth coordinates. 
The loss function $L_{root}$ is defined as follows:
\begin{equation}
L_{root} = \|\mathbf{R} - \mathbf{R}^*\|_1,
\end{equation}
where $*$ indicates the groundtruth.

\section{PoseNet}


\subsection{Model design}
The PoseNet estimates the root-relative 3D pose $\mathbf{P}^{rel}_{j}=(x_j,y_j,Z^{rel}_j)$ from a cropped human image. 
Many works have been presented for this topic~\cite{pavlakos2017coarse,sun2017compositional,zhou2017weaklysupervised,yang20183d,mehta2017monocular,martinez2017simple,sun2018integral}. 
Among them, we use the model of Sun~\etal~\cite{sun2018integral}, which is the current state-of-the-art method. 
This model consists of two parts. The first part is the backbone, which extracts a useful global feature from the cropped human image using ResNet~\cite{he2016deep}. 
Second, the pose estimation part takes a feature map from the backbone part and upsamples it using three consecutive deconvolutional layers with batch normalization layers~\cite{ioffe2015batch} and ReLU activation function. 
A 1-by-1 convolution is applied to the upsampled feature map to produce the 3D heatmaps for each joint. The soft-argmax operation is used to extract the 2D image coordinates $(x_j,y_j)$, and the root-relative depth values $Z^{rel}_j$.

\subsection{Loss function}
We train the PoseNet by minimizing the $L1$ distance between the estimated and groundtruth coordinates. The loss function $L_{pose}$ is defined as follows:
\begin{equation}
L_{pose} = \frac{1}{J} \sum_{j=1}^{J} \|\mathbf{P}_j^{rel} - \mathbf{P}_j^{rel*}\|_1,
\end{equation}
where $*$ indicates groundtruth.

\section{Implementation details}

Publicly released Mask R-CNN model~\cite{massa2018mrcnn} pre-trained on the COCO dataset~\cite{lin2014microsoft} is used for the DetectNet without fine-tuning on the human pose estimation datasets~\cite{ionescu2014human3,mehta2018single}. 
For the RootNet and PoseNet, PyTorch~\cite{paszke2017automatic} is used for implementation. Their backbone part is initialized with the publicly released ResNet-50~\cite{he2016deep} pre-trained on the ImageNet dataset~\cite{russakovsky2015imagenet}, and the weights of the remaining part are initialized by Gaussian distribution with $\sigma=0.001$. 
The weights are updated by the Adam optimizer~\cite{kingma2014adam} with a mini-batch size of 128. 
The initial learning rate is set to $1\times 10^{-3}$ and reduced by a factor of 10 at the 17th epoch. 
We use 256$\times$256 as the size of the input image of the RootNet and PoseNet. 
We perform data augmentation including rotation ($\pm$\ang{30}), horizontal flip, color jittering, and synthetic occlusion~\cite{zhong2017random} in training. 
Horizontal flip augmentation is performed in testing for the PoseNet following Sun~\etal~\cite{sun2018integral}. 
We train the RootNet and PoseNet for 20 epochs with four NVIDIA 1080 Ti GPUs, which took two days, respectively.

\section{Experiment}

\subsection{Dataset and evaluation metric}
\textbf{Human3.6M dataset.} 
Human3.6M dataset~\cite{ionescu2014human3} is the largest 3D single-person pose benchmark. 
It consists of 3.6 millions of video frames. 11 subjects performing 15 activities are captured from 4 camera viewpoints. 
The groundtruth 3D poses are obtained using a motion capture system. Two evaluation metrics are widely used. 
The first one is mean per joint position error (MPJPE)~\cite{ionescu2014human3}, which is calculated after aligning the human root of the estimated and groundtruth 3D poses. 
The second one is MPJPE after further alignment (\textit{i.e.}, Procrustes analysis (PA)~\cite{gower1975generalized}). 
This metric is called PA MPJPE. To evaluate the localization of the absolute 3D human root, we introduce the mean of the Euclidean distance between the estimated coordinates of the root $\mathbf{R}$ and the groundtruth $\mathbf{R}^{*}$, \textit{i.e.}, the mean of the root position error (MRPE), as a new metric:
\begin{equation}
MRPE=\frac{1}{N}\sum_{i=1}^{N}||\mathbf{R}^{(i)}-\mathbf{R}^{(i)*}||_{2},
\end{equation}
where superscript $i$ is the sample index, and $N$ denotes the total number of test samples.

\textbf{MuCo-3DHP and MuPoTS-3D datasets.}
These are the 3D multi-person pose estimation datasets proposed by Mehta~\etal~\cite{mehta2018single}.
The training set, MuCo-3DHP, is generated by compositing the existing MPI-INF-3DHP 3D single-person pose estimation dataset~\cite{mehta2017monocular}. 
The test set, MuPoTS-3D dataset, was captured at outdoors and it includes 20 real-world scenes with groundtruth 3D poses for up to three subjects. 
The groundtruth is obtained with a multi-view marker-less motion capture system. For evaluation, a 3D percentage of correct keypoints (3DPCK$_{rel}$) and area under 3DPCK curve from various thresholds (AUC$_{rel}$) is used after root alignment with groundtruth. 
It treats a joint's prediction as correct if it lies within a 15cm from the groundtruth joint location. 
We additionally define 3DPCK$_{abs}$ which is the 3DPCK without root alignment to evaluate the absolute camera-centered coordinates. 
To evaluate the localization of the absolute 3D human root, we use the average precision of 3D human root location ($AP_{25}^{root}$) which considers a prediction is correct when the Euclidean distance between the estimated and the groundtruth coordinates is smaller than 25cm.

\subsection{Experimental protocol}
\textbf{Human3.6M dataset.} 
Two experimental protocols are widely used. \textit{Protocol 1} uses six subjects (S1, S5, S6, S7, S8, S9) in training and S11 in testing. 
PA MPJPE is used as an evaluation metric. \textit{Protocol 2} uses five subjects (S1, S5, S6, S7, S8) in training and two subjects (S9, S11) in testing. 
MPJPE is used as an evaluation metric. 
We use every 5th and 64th frames in videos for training and testing, respectively following ~\cite{sun2017compositional,sun2018integral}. 
When training, besides the Human3.6M dataset, we used additional MPII 2D human pose estimation dataset~\cite{andriluka20142d} following ~\cite{zhou2017weaklysupervised,pavlakos2017coarse,sun2017compositional,sun2018integral}. 
Each mini-batch consists of half Human3.6M and half MPII data. For MPII data, the loss value of the $z$-axis becomes zero for both of the RootNet and PoseNet following Sun~\etal~\cite{sun2018integral}.

\begin{table}
\small
\centering
\setlength\tabcolsep{1.0pt}
\def\arraystretch{1.1}
\begin{tabular}{L{3.7cm}|C{1.5cm}C{1.5cm}C{1.5cm}}
\specialrule{.1em}{.05em}{.05em}
      Settings & MRPE & MPJPE & Time \\ \hline
Joint learning   & 138.2 & 116.7 & \textbf{0.132} \\ 
\textbf{Disjointed learning (Ours)}   & \textbf{120.0} & \textbf{57.3} & 0.141 \\ 
 \specialrule{.1em}{.05em}{.05em}
\end{tabular}
\vspace*{-3mm}
\caption{MRPE, MPJPE, and seconds per frame comparison between joint and disjointed learning on Human3.6M dataset.}
\vspace*{-5mm}
\label{table:mtl}
\end{table}

\begin{table}
\small
\centering
\setlength\tabcolsep{1.0pt}
\def\arraystretch{1.1}
\begin{tabular}{L{1.4cm}L{1.1cm}C{1.0cm}C{1.3cm}|C{1.3cm}C{1.7cm}}
\specialrule{.1em}{.05em}{.05em}
DetectNet & RootNet & $AP^{box}$ & $AP_{25}^{root}$ & AUC$_{rel}$ & 3DPCK$_{abs}$ \\ \hline
R-50 & $k$ & 43.8 & 5.2 & 39.2 & 9.6 \\
R-50 & Ours & 43.8 & 28.5 & 39.8 & 31.5 \\
X-101-32 & Ours & \textbf{45.0} & \textbf{31.0} & \textbf{39.8} & \textbf{31.5} \\ \hline
GT & Ours & 100.0 & 31.4 & 39.8 & 31.6 \\
GT & GT & 100.0 & 100.0 & 39.8 & 80.2 \\ \hline
\end{tabular}
\vspace*{-3mm}
\caption{Overall performance comparison for different DetectNet and RootNet settings on the MuPoTS-3D dataset.}
\vspace*{-3mm}
\label{table:effect_of_detectnet_rootnet}
\end{table}

\begin{table*}
\footnotesize
\centering
\setlength\tabcolsep{1.0pt}
\def\arraystretch{1.1}
\begin{tabular}{L{2.5cm}C{0.8cm}C{0.8cm}C{0.8cm}C{0.8cm}C{0.9cm}C{0.8cm}C{0.8cm}C{0.8cm}C{1.0cm}C{0.8cm}C{0.8cm}C{0.8cm}C{0.8cm}C{1.1cm}C{1.0cm}C{0.8cm}}
\specialrule{.1em}{.05em}{.05em}
Methods & Dir. & Dis. & Eat & Gre. & Phon. & Pose & Pur. & Sit & SitD. & Smo. & Phot. & Wait & Walk & WalkD. & WalkP. & Avg \\ \hline
\multicolumn{17}{l}{\textbf{\textit{With groundtruth information in inference time}}}  \\
Yasin~\cite{yasin2016dual} & 88.4 & 72.5 & 108.5 & 110.2 & 97.1 & 81.6 & 107.2 & 119.0 & 170.8 & 108.2 & 142.5 & 86.9 & 92.1 & 165.7 & 102.0 & 108.3 \\
Chen~\cite{chen20173d} & 71.6 & 66.6 & 74.7 & 79.1 & 70.1 & 67.6 & 89.3 & 90.7 & 195.6 & 83.5 & 93.3 & 71.2 & 55.7 & 85.9 & 62.5 & 82.7 \\
Moreno~\cite{moreno20173d} & 67.4 & 63.8 & 87.2 & 73.9 & 71.5 & 69.9 & 65.1 & 71.7 & 98.6 & 81.3 & 93.3 & 74.6 & 76.5 & 77.7 & 74.6 & 76.5 \\
Zhou~\cite{zhou2019monocap} & 47.9 & 48.8 & 52.7 & 55.0 & 56.8 & 49.0 & 45.5 & 60.8 & 81.1 & 53.7 & 65.5 & 51.6 & 50.4 & 54.8 & 55.9 & 55.3 \\
Martinez~\cite{martinez2017simple} & 39.5 & 43.2 & 46.4 & 47.0 & 51.0 & 41.4 & 40.6 & 56.5 & 69.4 & 49.2 & 56.0 & 45.0 & 38.0 & 49.5 & 43.1 & 47.7 \\
Sun~\cite{sun2017compositional} & 42.1 & 44.3 & 45.0 & 45.4 & 51.5 & 43.2 & 41.3 & 59.3 & 73.3 & 51.0 & 53.0 & 44.0 & 38.3 & 48.0 & 44.8 & 48.3 \\
Fang~\cite{fang2018learning} & 38.2 & 41.7 & 43.7 & 44.9 & 48.5 & 40.2 & 38.2 & 54.5 & 64.4 & 47.2 & 55.3 & 44.3 & 36.7 & 47.3 & 41.7 & 45.7 \\
Sun~\cite{sun2018integral} & 36.9 & 36.2 & 40.6 & 40.4 & 41.9 & 34.9 & 35.7 & 50.1 & 59.4 & 40.4 & 44.9 & 39.0 & 30.8 & 39.8 & 36.7 & 40.6 \\ 
\textbf{Ours (PoseNet)} & \textbf{31.0} & \textbf{30.6} & \textbf{39.9} & \textbf{35.5} & \textbf{34.8} & \textbf{30.2} & \textbf{32.1} & \textbf{35.0} & \textbf{43.8} & \textbf{35.7} & \textbf{37.6} & \textbf{30.1} & \textbf{24.6} & \textbf{35.7} & \textbf{29.3} & \textbf{34.0} \\ \hline
\multicolumn{17}{l}{\textbf{\textit{Without groundtruth information in inference time}}}  \\
Rogez~\cite{rogez2019lcr}$^*$ & - & - & - & - & - & - & - & - & - & - & - & - & - & - & - & 42.7 \\
\textbf{Ours (Full)} & \textbf{32.5} & \textbf{31.5} & \textbf{41.5} & \textbf{36.7} & \textbf{36.3} & \textbf{31.9} & \textbf{33.2} & \textbf{36.5} & \textbf{44.4} & \textbf{36.7} & \textbf{38.7} & \textbf{31.2} & \textbf{25.6} & \textbf{37.1} & \textbf{30.5} & \textbf{35.2} \\ \hline
\end{tabular}
\vspace*{-3mm}
\caption{PA MPJPE comparison with state-of-the-art methods on the Human3.6M dataset using Protocol 1. $*$ used extra synthetic data for training.}
\vspace*{-3mm}
\label{table:h36m_compare_protocol1}
\end{table*}

\begin{table*}
\footnotesize
\centering
\setlength\tabcolsep{1.0pt}
\def\arraystretch{1.1}
\begin{tabular}{L{2.5cm}C{0.8cm}C{0.8cm}C{0.8cm}C{0.8cm}C{0.9cm}C{0.8cm}C{0.8cm}C{0.8cm}C{1.0cm}C{0.8cm}C{0.8cm}C{0.8cm}C{0.8cm}C{1.1cm}C{1.0cm}C{0.8cm}}
\specialrule{.1em}{.05em}{.05em}
Methods & Dir. & Dis. & Eat & Gre. & Phon. & Pose & Pur. & Sit & SitD. & Smo. & Phot. & Wait & Walk & WalkD. & WalkP. & Avg \\ \hline
\multicolumn{17}{l}{\textbf{\textit{With groundtruth information in inference time}}}  \\
Chen~\cite{chen20173d} & 89.9 & 97.6 & 90.0 & 107.9 & 107.3 & 93.6 & 136.1 & 133.1 & 240.1 & 106.7 & 139.2 & 106.2 & 87.0 & 114.1 & 90.6 & 114.2 \\
Tome~\cite{tome2017lifting} & 65.0 & 73.5 & 76.8 & 86.4 & 86.3 & 68.9 & 74.8 & 110.2 & 173.9 & 85.0 & 110.7 & 85.8 & 71.4 & 86.3 & 73.1 & 88.4 \\
Moreno~\cite{moreno20173d} & 69.5 & 80.2 & 78.2 & 87.0 & 100.8 & 76.0 & 69.7 & 104.7 & 113.9 & 89.7 & 102.7 & 98.5 & 79.2 & 82.4 & 77.2 & 87.3 \\
Zhou~\cite{zhou2019monocap} & 68.7 & 74.8 & 67.8 & 76.4 & 76.3 & 84.0 & 70.2 & 88.0 & 113.8 & 78.0 & 98.4 & 90.1 & 62.6 & 75.1 & 73.6 & 79.9 \\
Jahangiri~\cite{jahangiri2017generating} & 74.4 & 66.7 & 67.9 & 75.2 & 77.3 & 70.6 & 64.5 & 95.6 & 127.3 & 79.6 & 79.1 & 73.4 & 67.4 & 71.8 & 72.8 & 77.6 \\
Mehta~\cite{mehta2017monocular} & 57.5 & 68.6 & 59.6 & 67.3 & 78.1 & 56.9 & 69.1 & 98.0 & 117.5 & 69.5 & 82.4 & 68.0 & 55.3 & 76.5 & 61.4 & 72.9 \\
Martinez~\cite{martinez2017simple} & 51.8 & 56.2 & 58.1 & 59.0 & 69.5 & 55.2 & 58.1 & 74.0 & 94.6 & 62.3 & 78.4 & 59.1 & 49.5 & 65.1 & 52.4 & 62.9 \\
Fang~\cite{fang2018learning} & 50.1 & 54.3 & 57.0 & 57.1 & 66.6 & 53.4 & 55.7 & 72.8 & 88.6 & 60.3 & 73.3 & 57.7 & 47.5 & 62.7 & 50.6 & 60.4 \\
Sun~\cite{sun2017compositional} & 52.8 & 54.8 & 54.2 & 54.3 & 61.8 & 53.1 & 53.6 & 71.7 & 86.7 & 61.5 & 67.2 & 53.4 & 47.1 & 61.6 & 63.4 & 59.1 \\
Sun~\cite{sun2018integral} & \textbf{47.5} & \textbf{47.7} & \textbf{49.5} & \textbf{50.2} & \textbf{51.4} & \textbf{43.8} & \textbf{46.4} & \textbf{58.9} & \textbf{65.7} & \textbf{49.4} & \textbf{55.8} & \textbf{47.8} & \textbf{38.9} & \textbf{49.0} & \textbf{43.8} & \textbf{49.6} \\ 
\textbf{Ours (PoseNet)} & 50.5 & 55.7 & 50.1 & 51.7 & 53.9 & 46.8 & 50.0 & 61.9 & 68.0 & 52.5 & 55.9 & 49.9 & 41.8 & 56.1 & 46.9 & 53.3 \\ \hline
\multicolumn{17}{l}{\textbf{\textit{Without groundtruth information in inference time}}}  \\
Rogez~\cite{rogez2017lcr} & 76.2 & 80.2 & 75.8 & 83.3 & 92.2 & 79.9 & 71.7 & 105.9 & 127.1 & 88.0 & 105.7 & 83.7 & 64.9 & 86.6 & 84.0 & 87.7 \\
Mehta~\cite{mehta2018single} & 58.2 & 67.3 & 61.2 & 65.7 & 75.8 & 62.2 & 64.6 & 82.0 & 93.0 & 68.8 & 84.5 & 65.1 & 57.6 & 72.0 & 63.6 & 69.9 \\
Rogez~\cite{rogez2019lcr}$^*$ & 55.9 & 60.0 & 64.5 & 56.3 & 67.4 & 71.8 & 55.1 & \textbf{55.3} & 84.8 & 90.7 & 67.9 & 57.5 & 47.8 & 63.3 & 54.6 & 63.5 \\
\textbf{Ours (Full)} & \textbf{51.5} & \textbf{56.8} & \textbf{51.2} & \textbf{52.2} & \textbf{55.2} & \textbf{47.7} & \textbf{50.9} & 63.3 & \textbf{69.9} & \textbf{54.2} & \textbf{57.4} & \textbf{50.4} & \textbf{42.5} & \textbf{57.5} & \textbf{47.7} & \textbf{54.4} \\ \hline
\end{tabular}
\vspace*{-3mm}
\caption{MPJPE comparison with state-of-the-art methods on the Human3.6M dataset using Protocol 2. $*$ used extra synthetic data for training.}
\vspace*{-3mm}
\label{table:h36m_compare_protocol2}
\end{table*}

\begin{table*}
\small
\centering
\setlength\tabcolsep{1.0pt}
\def\arraystretch{1.1}
\begin{tabular}{L{1.7cm}C{0.65cm}C{0.65cm}C{0.65cm}C{0.65cm}C{0.65cm}C{0.65cm}C{0.65cm}C{0.65cm}C{0.65cm}C{0.7cm}C{0.7cm}C{0.7cm}C{0.7cm}C{0.7cm}C{0.7cm}C{0.7cm}C{0.7cm}C{0.7cm}C{0.7cm}C{0.7cm}C{0.7cm}}
\specialrule{.1em}{.05em}{.05em}
Methods & S1 & S2 & S3 & S4 & S5 & S6 & S7 & S8 & S9 & S10 & S11 & S12 & S13 & S14 & S15 & S16 & S17 & S18 & S19 & S20 & Avg \\ \hline
\multicolumn{22}{l}{\textbf{\textit{Accuracy for all groundtruths}}}  \\
Rogez~\cite{rogez2017lcr} & 67.7 & 49.8 & 53.4 & 59.1 & 67.5 & 22.8 & 43.7 & 49.9 & 31.1 & 78.1 & 50.2 & 51.0 & 51.6 & 49.3 & 56.2 & 66.5 & 65.2 & 62.9 & 66.1 & 59.1 & 53.8 \\
Mehta~\cite{mehta2018single} & 81.0 & 60.9 & 64.4 & 63.0 & 69.1 & 30.3 & 65.0 & 59.6 & 64.1 & 83.9 & 68.0 & 68.6 & 62.3 & 59.2 & 70.1 & 80.0 & 79.6 & 67.3 & 66.6 & 67.2 & 66.0 \\
Rogez~\cite{rogez2019lcr}$^*$ & 87.3 & 61.9 & 67.9 & 74.6 & 78.8 & 48.9 & 58.3 & 59.7 & 78.1 & 89.5 & 69.2 & 73.8 & 66.2 & 56.0 & 74.1 & 82.1 & 78.1 & 72.6 & 73.1 & 61.0 & 70.6 \\
\textbf{Ours} & \textbf{94.4} & \textbf{77.5} & \textbf{79.0} & \textbf{81.9} & \textbf{85.3} & \textbf{72.8} & \textbf{81.9} & \textbf{75.7} & \textbf{90.2} & \textbf{90.4} & \textbf{79.2} & \textbf{79.9} & \textbf{75.1} & \textbf{72.7} & \textbf{81.1} & \textbf{89.9} & \textbf{89.6} & \textbf{81.8} & \textbf{81.7} & \textbf{76.2} & \textbf{81.8} \\ \hline
\multicolumn{22}{l}{\textbf{\textit{Accuracy only for matched groundtruths}}}  \\
Rogez~\cite{rogez2017lcr} & 69.1 & 67.3 & 54.6 & 61.7 & 74.5 & 25.2 & 48.4 & 63.3 & 69.0 & 78.1 & 53.8 & 52.2 & 60.5 & 60.9 & 59.1 & 70.5 & 76.0 & 70.0 & 77.1 & 81.4 & 62.4 \\
Mehta~\cite{mehta2018single} & 81.0 & 65.3 & 64.6 & 63.9 & 75.0 & 30.3 & 65.1 & 61.1 & 64.1 & 83.9 & 72.4 & 69.9 & 71.0 & 72.9 & 71.3 & 83.6 & 79.6 & 73.5 & 78.9 & \textbf{90.9} & 70.8 \\
Rogez~\cite{rogez2019lcr}$^*$ & 88.0 & 73.3 & 67.9 & 74.6 & 81.8 & 50.1 & 60.6 & 60.8 & 78.2 & 89.5 & 70.8 & 74.4 & 72.8 & 64.5 & 74.2 & 84.9 & 85.2 & 78.4 & 75.8 & 74.4 & 74.0 \\
\textbf{Ours} & \textbf{94.4} & \textbf{78.6} & \textbf{79.0} & \textbf{82.1} & \textbf{86.6} & \textbf{72.8} & \textbf{81.9} & \textbf{75.8} & \textbf{90.2} & \textbf{90.4} & \textbf{79.4} & \textbf{79.9} & \textbf{75.3} & \textbf{81.0} & \textbf{81.0} & \textbf{90.7} & \textbf{89.6} & \textbf{83.1} & \textbf{81.7} & 77.3 & \textbf{82.5} \\ \hline
\end{tabular}
\vspace*{-3mm}
\caption{Sequence-wise 3DPCK$_{rel}$ comparison with state-of-the-art methods on the MuPoTS-3D dataset. $*$ used extra synthetic data for training.}
\vspace*{-6mm}
\label{table:mupots_compare_sequence}
\end{table*}

\begin{table}
\small
\centering
\setlength\tabcolsep{1.0pt}
\def\arraystretch{1.1}
\begin{tabular}{L{1.5cm}C{0.6cm}C{0.7cm}C{0.7cm}C{0.7cm}C{0.7cm}C{0.7cm}C{0.7cm}C{0.7cm}C{0.7cm}}
\specialrule{.1em}{.05em}{.05em}
Methods & Hd. & Nck. & Sho. & Elb. & Wri. & Hip & Kn. & Ank. & Avg \\ \hline
Rogez~\cite{rogez2017lcr} & 49.4 & 67.4 & 57.1 & 51.4 & 41.3 & 84.6 & 56.3 & 36.3 & 53.8 \\
Mehta~\cite{mehta2018single} & 62.1 & 81.2 & 77.9 & 57.7 & 47.2 & \textbf{97.3} & 66.3 & 47.6 & 66.0 \\
\textbf{Ours} & \textbf{79.1} & \textbf{92.6} & \textbf{85.1} & \textbf{79.4} & \textbf{67.0} & 96.6 & \textbf{85.7} & \textbf{73.1} & \textbf{81.8} \\ \hline
\end{tabular}
\vspace*{-3mm}
\caption{Joint-wise 3DPCK$_{rel}$ comparison with state-of-the-art methods on the MuPoTS-3D dataset. All groundtruths are used for evaluation.}
\vspace*{-3mm}
\label{table:mupots_compare_joint}
\end{table}

\textbf{MuCo-3DHP and MuPoTS-3D datasets.}
Following the previous protocol, we composite 400K frames of which half are background augmented. 
For augmentation, we use images from the COCO dataset~\cite{lin2014microsoft} except for images with humans. 
We use an additional COCO 2D human keypoint detection dataset~\cite{lin2014microsoft} when training our models on the MuCo-3DHP dataset following Mehta~\etal~\cite{mehta2018single}. 
Each mini-batch consists of half MuCo-3DHP and half COCO data. For COCO data, loss value of $z$-axis becomes zero for both of the RootNet and PoseNet following Sun~\etal~\cite{sun2018integral}.

\subsection{Ablation study}
In this study, we show how each component of our proposed framework affects the 3D multi-person pose estimation accuracy. 
To evaluate the performance of the DetectNet, we use the average precision of bounding box ($AP^{box}$) following metrics of the COCO object detection benchmark~\cite{lin2014microsoft}. 

\textbf{Disjointed pipeline.}
To demonstrate the effectiveness of the disjointed pipeline (\textit{i.e.}, separated DetectNet, RootNet, and PoseNet), we compare MRPE, MPJPE, and running time of joint and disjointed learning of the RootNet and PoseNet in Table~\ref{table:mtl}. 
The running time includes DetectNet and is measured using a single TitanX Maxwell GPU. 
For the joint learning, we combine the RootNet and PoseNet into a single model which shares backbone part (\textit{i.e.}, ResNet~\cite{he2016deep}). 
The image feature from the backbone is fed to each branch of RootNet and PoseNet in a parallel way. Compared with the joint learning, our disjointed learning gives lower error under a similar running time. 
We believe that this is because each task of RootNet and PoseNet is not highly correlated so that jointly training all tasks can make training harder, resulting in lower accuracy.

\textbf{Effect of the DetectNet.}
To show how the performance of the human detection affects the accuracy of the final 3D human root localization and 3D multi-person pose estimation, we compare AP$_{25}^{root}$, AUC$_{rel}$, and 3DPCK$_{abs}$ using the DetectNet in various backbones (\textit{i.e.}, ResNet-50~\cite{he2016deep}, ResNeXt-101-32~\cite{xie2017aggregated}) and groundtruth box in the second, third, and fourth row of Table~\ref{table:effect_of_detectnet_rootnet}, respectively. 
The table shows that based on the same RootNet (\textit{i.e.}, Ours), better human detection model improves both of the 3D human root localization and 3D multi-person pose estimation performance. However, the groundtruth box does not improve overall accuracy considerably compared with other DetectNet models. 
Therefore, we have sufficient reasons to believe that the given boxes cover most of the person instances with such a high detection AP. 
We can also conclude that the bounding box estimation accuracy does not have a large impact on the 3D multi-person pose estimation accuracy.

\textbf{Effect of the RootNet.}
To show how the performance of the 3D human root localization affects the accuracy of the 3D multi-person pose estimation, we compare AUC$_{rel}$ and 3DPCK$_{abs}$ using various RootNet settings in Table~\ref{table:effect_of_detectnet_rootnet}. 
The first and second rows show that based on the same DetectNet (\textit{i.e.}, R-50), our RootNet exhibits significantly higher AP$_{25}^{root}$ and 3DPCK$_{abs}$ compared with the setting in which $k$ is directly utilized as a depth value. We use the $x$ and $y$ of the RootNet when the $k$ is used as a depth value. 
This result demonstrates that the RootNet successfully corrects the $k$ value. The fourth and last rows show that the groundtruth human root provides similar AUC$_{rel}$, but significantly higher 3DPCK$_{abs}$ compared with our RootNet. 
This finding shows that better human root localization is required to achieve more accurate absolute 3D multi-person pose estimation results.

\textbf{Effect of the PoseNet.}
All settings in Table~\ref{table:effect_of_detectnet_rootnet} provides similar AUC$_{rel}$. Especially, the first and last rows of the table show that using groundtruth box and human root does not provide significantly higher AUC$_{rel}$. 
As the results in the table are based on the same PoseNet, we can conclude that AUC$_{rel}$, which is an evaluation of the root-relative 3D human pose estimation highly depends on the accuracy of the PoseNet.

\subsection{Comparison with state-of-the-art methods}
\textbf{Human3.6M dataset.}
We compare our proposed system with the state-of-the-art 3D human pose estimation methods on the Human3.6M dataset~\cite{ionescu2014human3} in Tables~\ref{table:h36m_compare_protocol1} and ~\ref{table:h36m_compare_protocol2}. 
As most of the previous methods use the groundtruth information (\textit{i.e.}, bounding boxes or 3D root locations) in inference time, we report the performance of the PoseNet using the groundtruth 3D root location. 
Note that our full model does not require any groundtruth information in inference time. The tables show that our method achieves comparable performance despite not using any groundtruth information in inference time. 
Moreover, it significantly outperforms previous 3D multi-person pose estimation methods~\cite{lin2014microsoft,mehta2018single}. 

\textbf{MuCo-3DHP and MuPoTS-3D datasets.}
We compare our proposed system with the state-of-the-art 3D multi-person pose estimation methods on the MuPoTS-3D dataset~\cite{mehta2018single} in Tables~\ref{table:mupots_compare_sequence} and ~\ref{table:mupots_compare_joint}. 
The proposed system significantly outperforms them in most of the test sequences and joints. 


\section{Discussion}

Although our proposed method outperforms previous 3D multi-person pose estimation methods by a large margin, room for improvement is substantial. As shown in Table~\ref{table:effect_of_detectnet_rootnet}, using the groundtruth 3D root location brings significant 3DPCK$_{abs}$ improvement. Recent advances in depth map estimation from a single RGB image~\cite{fu2018deep,li2018deep} can give a clue for improving the 3D human root localization model. 

Our framework can also be used in applications other than 3D multi-person pose estimation. For example, recent methods for 3D human mesh model reconstruction~\cite{bogo2016keep,kanazawa2018end,joo2018total} reconstruct full 3D mesh model from a single person. Joo~\etal~\cite{joo2018total} utilized 2D multi-view input for 3D multi-person mesh model reconstruction. In our framework, if the PoseNet is replaced with existing human mesh reconstruction model~\cite{bogo2016keep,kanazawa2018end,joo2018total}, 3D multi-person mesh model reconstruction can be performed from \emph{a single RGB image}. This shows our framework can be applied to many 3D instance-aware vision tasks which take a single RGB image as an input.


\vspace*{-3mm}

\section{Conclusion}
We propose a novel and general framework for 3D multi-person pose estimation from a single RGB image. Our framework consists of human detection, 3D human root localization, and root-relative 3D single-person pose estimation models. Since any existing human detection and 3D single-person pose estimation models can be plugged into our framework, it is very flexible and easy to use. The proposed system outperforms previous 3D multi-person pose estimation methods by a large margin and achieves comparable performance with 3D single-person pose estimation methods without any groundtruth information while they use it in inference time. To the best of our knowledge, this work is the first to propose a fully learning-based camera distance-aware top-down approach whose components are compatible with most of the previous human detection and 3D human pose estimation models. We hope that this study provides a new basis for 3D multi-person pose estimation, which has only barely been explored.

\section*{Acknowledgments}
This work was partially supported by the Visual Turing Test project (IITP-2017-0-01780) from the Ministry of Science and ICT of Korea.

\clearpage

\begin{center}
\textbf{\large Supplementary Material of \enquote{Camera Distance-aware Top-down Approach for 3D Multi-person Pose Estimation from a Single RGB Image}}
\end{center}

\setcounter{section}{0}

In this supplementary material, we present more experimental results that could not be included in the main manuscript due to the lack of space.

\section{Derivation of Equation 1}

\begin{figure}[t]
\begin{center}
   \includegraphics[width=1.0\linewidth]{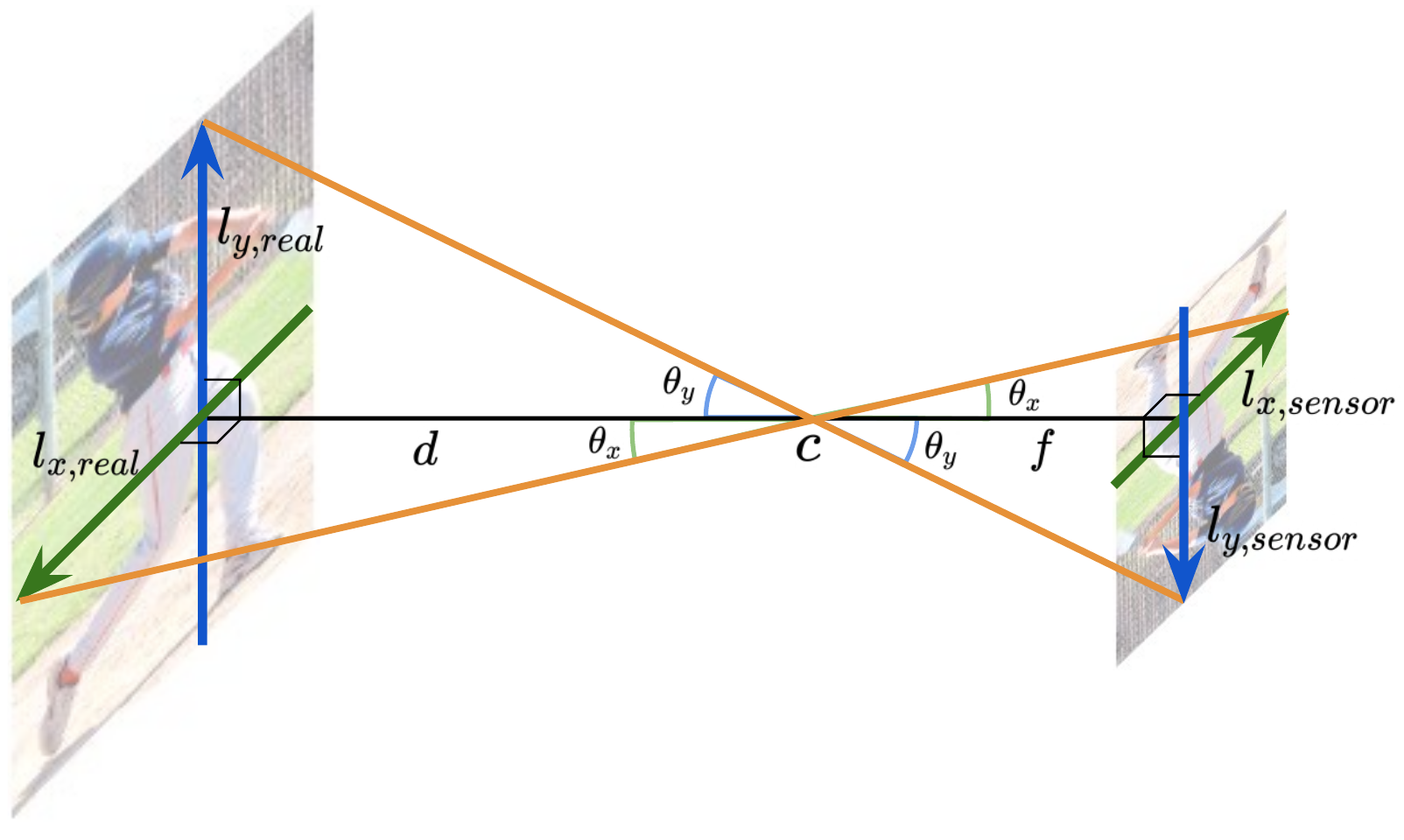}
\end{center}
\vspace*{-5mm}
   \caption{Visualization of a pinhole camera model.}
\vspace*{-3mm}
\label{fig:camera_model}
\end{figure}

We provide a derivation of Equation 1 of the main manuscript with reference to Figure~\ref{fig:camera_model} ,which shows a pinhole camera model. The green and blue arrows represent the human root joint centered $x$ and $y$-axes, respectively. The yellow lines show rays, and $c$ is the hole. $d$, $f$, and $l_{sensor}$ are distance between camera and the human root joint ($mm$), focal length ($mm$), and the length of human on the image sensor ($mm$), respectively. 

According to the definition of $\tan$,

\begin{equation}
\tan \theta_x = \frac{0.5l_{x,real}}{d} = \frac{0.5l_{x,sensor}}{f}, \nonumber
\end{equation}

Let $p_x$ be per pixel distance factor in $x$-axis. Then,
\begin{equation}
d = f \frac{l_{x,real}}{l_{x,sensor}} = f p_x \frac{l_{x,real}}{l_{x,sensor} p_x} = \alpha_x \frac{l_{x,real}}{l_{x,img}}, \nonumber
\end{equation}

Above equations are also valid in $y$-axis. Therefore, 
\begin{equation}
d = f \frac{l_{y,real}}{l_{y,sensor}} = f p_y \frac{l_{y,real}}{l_{y,sensor} p_y} = \alpha_y \frac{l_{y,real}}{l_{y,img}}, \nonumber
\end{equation}

Finally,
\begin{equation}
d = \sqrt{\alpha_x \alpha_y \frac{l_{x,real}}{l_{x,img}} \frac{l_{y,real}}{l_{y,img}}} = \sqrt{\alpha_x \alpha_y \frac{A_{real}}{A_{img}}}. \nonumber
\end{equation}

\section{Comparison of 3D human root localization with previous approaches}

We compare previous absolute 3D human root localization methods~\cite{rogez2017lcr,mehta2017monocular} with the proposed RootNet on the Human3.6M dataset~\cite{ionescu2014human3} based on protocol 2.

\begin{table}
\centering
\setlength\tabcolsep{1.0pt}
\def\arraystretch{1.1}
\begin{tabular}{L{2.7cm}|C{1.3cm}C{1.3cm}C{1.3cm}C{1.3cm}}
\specialrule{.1em}{.05em}{.05em}
Methods & MRPE & MRPE$_x$ & MRPE$_y$ & MRPE$_z$ \\ \hline
Baseline~\cite{rogez2017lcr,mehta2017monocular} & 267.8 & 27.5 & 28.3 & 261.9 \\
W/o limb joints & 226.2 & 24.5 & 24.9 & 220.2 \\
RANSAC & 213.1 & 24.3 & 24.3 & 207.1 \\
\textbf{RootNet (Ours)} & \textbf{120.0} & \textbf{23.3} & \textbf{23.0} & \textbf{108.1} \\ \hline
\end{tabular}
\vspace*{-3mm}
\caption{MRPE comparisons between previous distance minimization-based approaches~\cite{rogez2017lcr,mehta2017monocular} and our RootNet on the Human3.6M dataset. MRPE$_x$, MRPE$_y$, and MRPE$_z$ represent the mean of the errors in the $x$, $y$, and $z$ axes, respectively.}
\label{table:compare_rootnet}
\end{table}

\begin{table}
\centering
\setlength\tabcolsep{1.0pt}
\def\arraystretch{1.1}
\begin{tabular}{C{2.0cm}C{2.0cm}C{2.1cm}C{2.0cm}}
\specialrule{.1em}{.05em}{.05em}
      DetectNet & RootNet & PoseNet & \textbf{Total}  \\ \hline
0.120    & 0.010 & 0.011 & \textbf{0.141} \\ 
 \specialrule{.1em}{.05em}{.05em}
\end{tabular}
\vspace*{-3mm}
\caption{Seconds per frame for each component of our framework.}
\label{table:running_time}
\end{table}

\begin{table*}
\small
\centering
\setlength\tabcolsep{1.0pt}
\def\arraystretch{1.1}
\begin{tabular}{L{1.7cm}C{0.65cm}C{0.65cm}C{0.65cm}C{0.65cm}C{0.65cm}C{0.65cm}C{0.65cm}C{0.65cm}C{0.65cm}C{0.7cm}C{0.7cm}C{0.7cm}C{0.7cm}C{0.7cm}C{0.7cm}C{0.7cm}C{0.7cm}C{0.7cm}C{0.7cm}C{0.7cm}C{0.7cm}}
\specialrule{.1em}{.05em}{.05em}
Methods & S1 & S2 & S3 & S4 & S5 & S6 & S7 & S8 & S9 & S10 & S11 & S12 & S13 & S14 & S15 & S16 & S17 & S18 & S19 & S20 & Avg \\ \hline
\multicolumn{22}{l}{\textbf{\textit{Accuracy for all groundtruths}}}  \\
Ours & 59.5 & 44.7 & 51.4 & 46.0 & 52.2 & 27.4 & 23.7 & 26.4 & 39.1 & 23.6 & 18.3 & 14.9 & 38.2 & 26.5 & 36.8 & 23.4 & 14.4 & 19.7 & 18.8 & 25.1 & 31.5 \\ \hline
\multicolumn{22}{l}{\textbf{\textit{Accuracy only for matched groundtruths}}}  \\
Ours & 59.5 & 45.3 & 51.4 & 46.2 & 53.0 & 27.4 & 23.7 & 26.4 & 39.1 & 23.6 & 18.3 & 14.9 & 38.2 & 29.5 & 36.8 & 23.6 & 14.4 & 20.0 & 18.8 & 25.4 & 31.8 \\ \hline
\end{tabular}
\vspace*{-3mm}
\caption{Sequence-wise 3DPCK$_{abs}$ on the MuPoTS-3D dataset.}
\vspace*{-3mm}
\label{table:mupots_abs_sequence}
\end{table*}

\begin{table}
\small
\centering
\setlength\tabcolsep{1.0pt}
\def\arraystretch{1.1}
\begin{tabular}{L{1.5cm}C{0.6cm}C{0.7cm}C{0.7cm}C{0.7cm}C{0.7cm}C{0.7cm}C{0.7cm}C{0.7cm}C{0.7cm}}
\specialrule{.1em}{.05em}{.05em}
Methods & Hd. & Nck. & Sho. & Elb. & Wri. & Hip & Kn. & Ank. & Avg \\ \hline
Ours & 37.3 & 35.3 & 33.7 & 33.8 & 30.4 & 30.3 & 31.0 & 25.0 & 31.5 \\ \hline
\end{tabular}
\vspace*{-3mm}
\caption{Joint-wise 3DPCK$_{abs}$ on the MuPoTS-3D dataset. All groundtruths are used for evaluation.}
\vspace*{-3mm}
\label{table:mupots_abs_joint}
\end{table}

Previous approaches~\cite{rogez2017lcr,mehta2017monocular} simultaneously estimate 2D image coordinates and 3D camera-centered root-relative coordinates of keypoints. Then, absolute camera-centered coordinates of the human root are obtained by minimizing the distance between 2D predictions and projected 3D predictions. For optimization, linear least-squares formulation is used. To measure the errors of their method, we implemented and used ResNet-152-based model of Sun~\etal~\cite{sun2018integral} as a 2D pose estimator and model of Martinez~\etal~\cite{martinez2017simple} as a 3D pose estimator, which are state-of-the-art methods. In addition, to minimize the effect of outliers in 3D-to-2D fitting, we excluded limb joints when fitting. Also, we performed RANSAC with a various number of joints to get optimal joint set for fitting instead of using heuristically selected joint set.

Table~\ref{table:compare_rootnet} shows our RootNet significantly outperforms previous approaches. Furthermore, the RootNet can be designed independently of the PoseNet, giving design flexibility to both models. In contrast, the previous 3D root localization methods~\cite{rogez2017lcr,mehta2017monocular} require both of 2D and 3D predictions for the root localization, which results in lack of generalizability.

\section{Running time of the proposed framework}
In Table~\ref{table:running_time}, we report seconds per frame for each component of our framework. The running time is measured using a single TitanX Maxwell GPU. As the table shows, most of the running time is consumed by DetectNet. It is hard to directly compare running time with previous works~\cite{rogez2017lcr,mehta2017monocular} because they did not report it. However, we guess that there would be no big difference because models of ~\cite{rogez2017lcr} and ~\cite{mehta2017monocular} are similar with ~\cite{ren2015faster} and ~\cite{cao2016realtime} whose speed is 0.2 and 0.11 seconds per frame, respectively.

\section{Absolute 3D multi-person pose estimation errors}
For the continual study of the 3D multi-person pose estimation, we report 3DPCK$_{abs}$ in Table~\ref{table:mupots_abs_sequence} and ~\ref{table:mupots_abs_joint}. As previous works~\cite{lin2014microsoft,mehta2018single} did not report 3DPCK$_{abs}$, we only report our result.

\section{Qualitative results}
\begin{figure*}
\begin{center}
\includegraphics[width=0.9\linewidth]{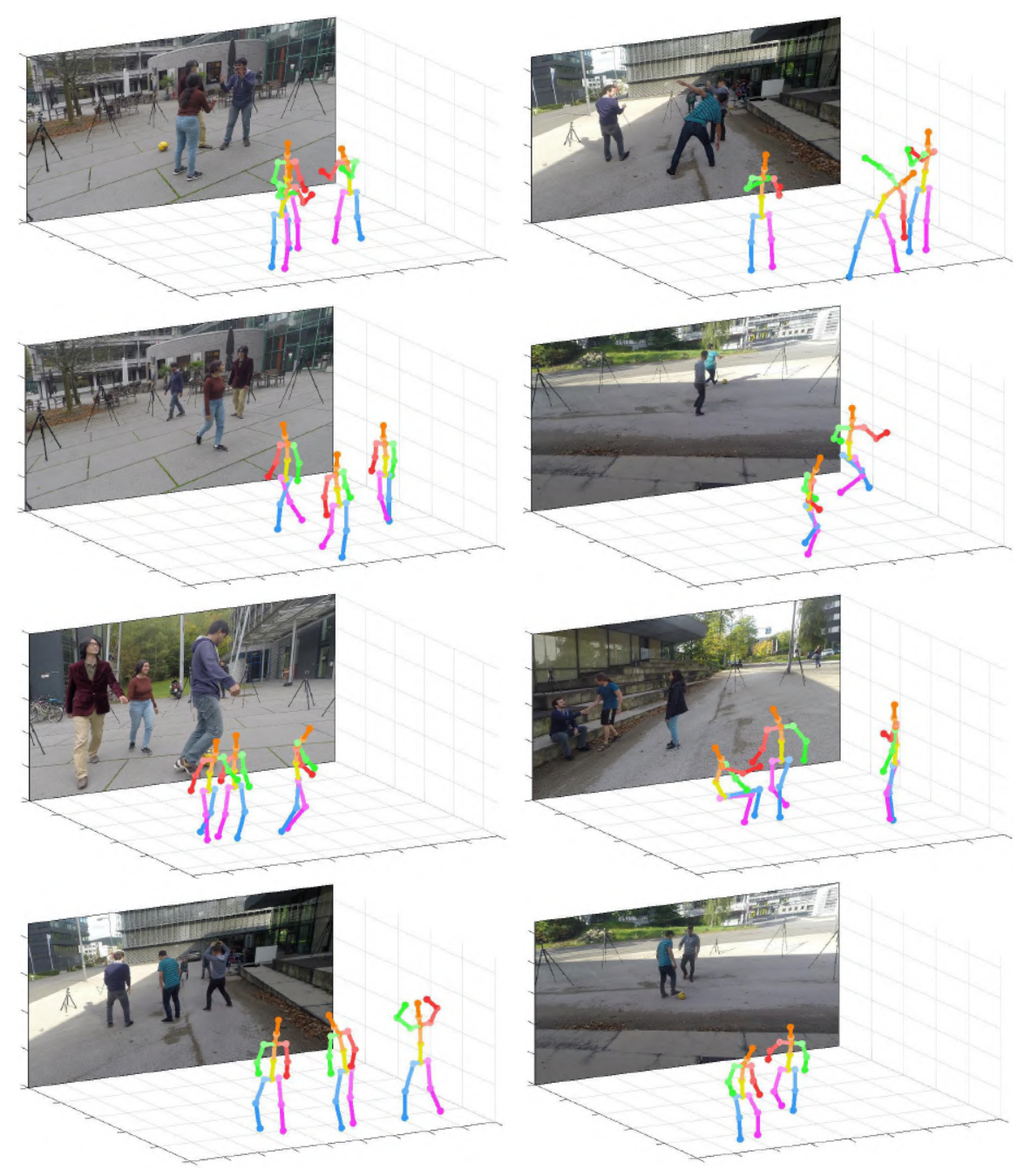}
\end{center}
\vspace*{-5mm}
   \caption{Qualitative results of applying our method on the MuPoTS-3D dataset~\cite{mehta2018single}.}
\vspace*{-3mm}
\label{fig:qualitative_mupots}
\end{figure*}

\begin{figure*}
\begin{center}
\includegraphics[width=0.9\linewidth]{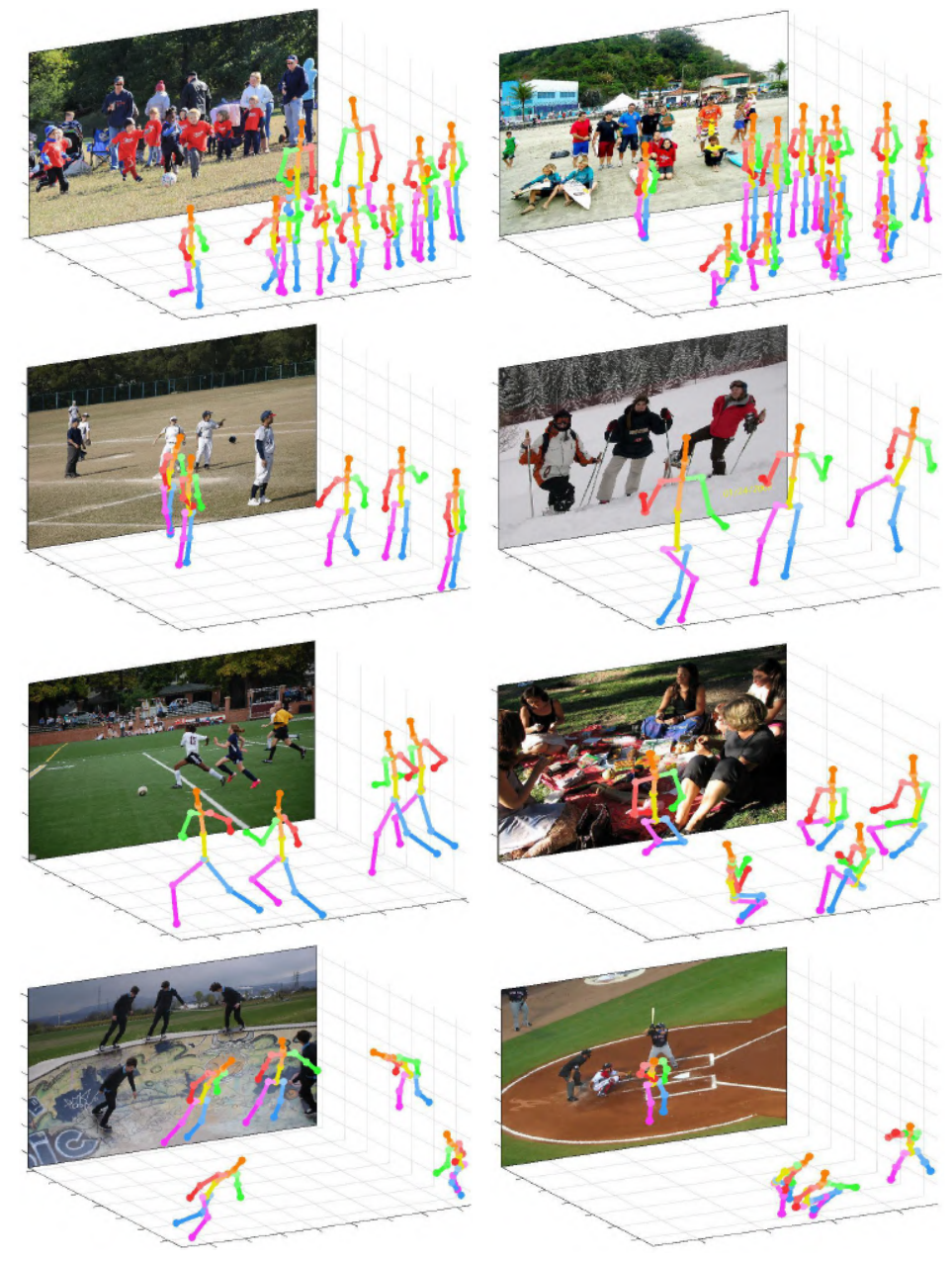}
\end{center}
\vspace*{-5mm}
   \caption{Qualitative results of applying our method on the COCO 2017~\cite{lin2014microsoft} validation set.}
\vspace*{-3mm}
\label{fig:qualitative_coco}
\end{figure*}

Figures~\ref{fig:qualitative_mupots} and ~\ref{fig:qualitative_coco} show qualitative results of our 3D multi-person pose estimation framework on the MuPoTS-3D~\cite{mehta2018single} and COCO~\cite{lin2014microsoft} datasets, respectively. Note that COCO dataset consists of \emph{in-the-wild} images which are hardly included in the 3D human pose estimation training sets~\cite{ionescu2014human3,mehta2018single}.

\clearpage

{\small
\bibliographystyle{ieee_fullname}
\bibliography{egpaper_final}
}

\end{document}